\RequirePackage[svgnames,table]{xcolor}
\documentclass[10pt,letterpaper,twocolumn]{yalearxiv}
\usepackage[numbers,sort&compress]{natbib}
\usepackage{booktabs,tabularx,makecell}
\usepackage{mathtools}
\usepackage{microtype}
\usepackage{doi}
\usepackage{balance}
\usepackage{fontawesome5}   
\hypersetup{
  pdftitle={Beyond the Previous Layer: Residual Predictive Structure in Sparse MoE Routing},
  pdfauthor={Hao Li; Yasuyuki Tahara; Yuichi Sei},
  pdfsubject={Held-out prediction of sparse MoE routing from expert-selection history},
  pdfkeywords={mixture of experts, routing history, nonlinear probes, residual prediction}
}
\setcitestyle{square}

\newcommand{\rsq}{R^2}

\newcommand{\metric}[1]{\csname val#1\endcsname}
\renewcommand{\HorRule}{\color{black}\rule{\linewidth}{1pt}}

\expandafter\def\csname valEpdPathContent\endcsname{0.66728}
\expandafter\def\csname valEipcEightA\endcsname{0.10923}
\expandafter\def\csname valEipcTwelveA\endcsname{0.17506}
\expandafter\def\csname valEipcMeanA\endcsname{0.14215}
\expandafter\def\csname valEpdId\endcsname{0.66969}
\expandafter\def\csname valEpdContent\endcsname{0.06732}
\expandafter\def\csname valEpdFirst\endcsname{0.14253}
\expandafter\def\csname valEpdLast\endcsname{0.59927}
\expandafter\def\csname valRmoOne\endcsname{0.59879}
\expandafter\def\csname valRmoEleven\endcsname{0.66544}
\expandafter\def\csname valRmoStepEleven\endcsname{0.00279}
\expandafter\def\csname valRmcTwelveDelta\endcsname{0.14275}
\expandafter\def\csname valRmcTwentyDelta\endcsname{0.20528}
\expandafter\def\csname valNhdTwelveA\endcsname{0.00139}
\expandafter\def\csname valNhdTwelveB\endcsname{0.17137}
\expandafter\def\csname valNhdTwelveMatched\endcsname{0.16440}
\expandafter\def\csname valNhdTwelveResidual\endcsname{0.20549}
\expandafter\def\csname valNhdTwelvePermutation\endcsname{-0.33325}
\expandafter\def\csname valNhdTwentyA\endcsname{0.00936}
\expandafter\def\csname valNhdTwentyB\endcsname{0.21861}
\expandafter\def\csname valNhdTwentyMatched\endcsname{0.21912}
\expandafter\def\csname valNhdTwentyResidual\endcsname{0.23556}
\expandafter\def\csname valNhdTwentyPermutation\endcsname{-0.32149}

\title{Beyond the Previous Layer:\\Residual Predictive Structure in Sparse MoE Routing}
\runningtitle{Beyond the Previous Layer: Residual Predictive Structure in Sparse MoE Routing}
\author[1]{Hao Li}
\author[1]{Yasuyuki Tahara}
\author[1]{Yuichi Sei\textsuperscript{*}}
\affil[1]{Department of Informatics, Graduate School of Informatics and Engineering,\\
The University of Electro-Communications, Tokyo, Japan\protect\\
\faGithub~\textbf{Source Code:} \href{https://github.com/withfanta/moe-routing-dynamics/tree/main/code}{\texttt{https://github.com/withfanta/moe-routing-dynamics/tree/main/code}}}
\paperurl{https://github.com/withfanta/moe-routing-dynamics/tree/main/code}
\begin{document}
\begin{abstract}
Sparse mixture-of-experts models route each token through a sequence of expert selections. We ask whether the immediately preceding selection adequately summarizes this trajectory for predicting the next router. Using frozen OLMoE and JetMoE models, we measure the held-out predictive gain from earlier expert selections while retaining the most recent selection as a common baseline. In OLMoE, extending the history from one to eleven layers raises router-logit $\rsq$ from 0.59879 to 0.66544. A preregistered JetMoE replication yields four-layer gains of 0.14275 and 0.20528 at two target depths, with paired bootstrap intervals above zero. These gains survive nonlinear decoding: adding history to a small multilayer perceptron improves $\rsq$ by 0.17137 and 0.21861, whereas nonlinear decoding of the recent state alone adds 0.00139 and 0.00936 over a linear probe. Parameter-matched controls preserve the advantage, and cross-fitted history residuals predict target residuals with $\rsq$ of 0.20549 and 0.23556. These findings identify residual predictive structure in expert-selection trajectories beyond adjacent-layer persistence.
\end{abstract}
\maketitle
\begingroup
\renewcommand{\thefootnote}{*}
\footnotetext[1]{\raggedright Corresponding author: Yuichi Sei.
E-mails: \href{mailto:r2540015@gl.cc.uec.ac.jp}{r2540015@gl.cc.uec.ac.jp};
\href{mailto:tahara@uec.ac.jp}{tahara@uec.ac.jp};
\href{mailto:seiuny@uec.ac.jp}{seiuny@uec.ac.jp}.\\
ORCID: Hao Li, \href{https://orcid.org/0009-0004-2623-286X}{0009-0004-2623-286X};
Yasuyuki Tahara, \href{https://orcid.org/0000-0002-1939-4455}{0000-0002-1939-4455};
Yuichi Sei, \href{https://orcid.org/0000-0002-2552-6717}{0000-0002-2552-6717}.}
\endgroup

\section{Introduction}
A token traversing a sparse mixture-of-experts (MoE) model acquires an expert-selection trajectory across depth. The immediately preceding selection offers a natural summary of that trajectory. How much predictive structure remains in earlier selections once this recent state is already available? Answering this question distinguishes adjacent-layer persistence from the additional value of a deeper routing history.

Cross-layer routing structure has motivated both analysis and architecture design. Read-ME decouples routing from the backbone to support pre-gating~\citep{readme}; RMoE propagates a recurrent routing state across layers~\citep{rmoe}; and PathMoE shares router parameters within consecutive blocks~\citep{pathmoe}. Labzin et al.\ align router-control subspaces and study shared linear dynamics~\citep{geometry}. Expert paths are also recorded and replayed during MoE reinforcement learning~\citep{esrl}, while HeRo uses routing history for dynamic layer skipping~\citep{hero}. Our question concerns the incremental predictive value of older selections in existing pretrained sparse MoEs, with the immediately preceding selection explicitly included in both sides of the comparison.

The evidence follows a sequence of increasingly specific questions. Selected-expert provenance predicts later routing better than fused expert outputs. Decomposing that provenance identifies expert selection as the dominant accessible signal. Holding the most recent selection fixed then reveals an additional history gain in OLMoE~\citep{olmoe}, which a preregistered JetMoE~\citep{jetmoe} experiment replicates at two depths. Finally, nonlinear decoders and cross-fitted residual prediction test whether the gain arises simply because a linear probe cannot decode the recent state.

We make three contributions: (1) an evaluation of routing-history value beyond an explicit previous-layer baseline; (2) evidence across two pretrained MoE architectures, including a preregistered replication; and (3) nonlinear and parameter-matched controls, together with residual prediction, that preserve the history effect. The resulting measurement provides a concrete target for understanding which aspects of a routing trajectory a local selection state summarizes.

\begin{table*}[t]
\centering\small
\caption{\textbf{Where the predictive signal comes from.} Held-out router-logit $R^2$ in OLMoE. EIPC retains selected contributions in identity slots; EPD separates selection identity from contribution content. EIPC and EPD use different samples and compression budgets, so comparisons are within each panel. The EPD single-layer curve uses 16 PCA components per representation.}
\label{tab:provenance}
\begin{minipage}[t]{0.49\textwidth}
\vspace{0pt}
\centering
\begin{tabular}{lrrr}
\toprule
\multicolumn{4}{l}{\textbf{EIPC: provenance (64 components)}}\\
Target & Fused & Identity & Shuffled\\
\midrule
L8 & 0.19016 & 0.29939 & -0.00149 \\
L12 & 0.28705 & 0.46211 & 0.00396 \\
\midrule
Target & Id.$-$Fused & Id.$-$Shuf. & \\
L8 & 0.10923 & 0.30088 &  \\
L12 & 0.17506 & 0.45815 &  \\
Mean & 0.14215 & 0.37952 &  \\
\bottomrule
\end{tabular}
\par\medskip
\begin{tabular}{lr}
\toprule
\multicolumn{2}{l}{\textbf{EPD: L12 (32 components)}}\\
Representation & $R^2$\\
\midrule
Fused & 0.20114 \\
Selection path & 0.66969 \\
Content by router rank & 0.06732 \\
Full provenance & 0.34007 \\
Path + content (post-hoc) & 0.66728 \\
\bottomrule
\end{tabular}
\end{minipage}\hfill
\begin{minipage}[t]{0.47\textwidth}
\vspace{0pt}
\centering
\begin{tabular}{rrr}
\toprule
\multicolumn{3}{l}{\textbf{EPD: single-layer prediction of L12}}\\
Source layer & Selection path & Content\\
\midrule
1 & 0.14253 & 0.00914 \\
2 & 0.25597 & -0.00721 \\
3 & 0.35859 & -0.00605 \\
4 & 0.41664 & 0.00556 \\
5 & 0.40384 & -0.01048 \\
6 & 0.44991 & -0.00814 \\
7 & 0.44797 & -0.00657 \\
8 & 0.49381 & 0.02139 \\
9 & 0.53850 & 0.00420 \\
10 & 0.57216 & 0.00700 \\
11 & 0.59927 & 0.01017 \\
\midrule
Mean & 0.42538 & 0.00173 \\
\bottomrule
\end{tabular}
\end{minipage}
\end{table*}

\section{Measuring Historical Value}
\subsection{State, target, and comparison}
For a fixed token, let $S_l$ be the binary vector indicating the experts selected at layer $l$, and let $g_l$ be that layer's native router-logit vector. Layer numbers are one-based. History here runs across model depth for the same token. It does not denote earlier tokens in the sequence.

A probe $F_k$ predicts $g_l$ from the $k$ immediately preceding selection states, denoted $S_{l-k:l-1}$. We evaluate
\begin{equation}
\mathcal{R}_l(k)=\rsq_{\mathrm{TEST}}\!\left(F_k(S_{l-k:l-1}),g_l\right),
\end{equation}
and measure the historical gain
\begin{equation}
\Delta_l(k)=\mathcal{R}_l(k)-\mathcal{R}_l(1).
\end{equation}
The one-layer baseline measures adjacent-layer predictability. A positive $\Delta_l(k)$ measures the improvement available to the tested probe when it also receives older selections. Throughout, $\rsq$ is the uniform average of the coordinate-wise scores. Target logits are centered within each sample and standardized using FIT statistics.

\subsection{Frozen evaluation and decoder controls}
All backbones are frozen, non-quantized, and evaluated in FP16: \texttt{OLMoE-1B-7B-0125} (revision \texttt{9b0c1aa8}) and \texttt{jetmoe-8b} (\texttt{d8fd02cc}). OLMoE selects eight of 64 MLP experts; JetMoE selects two of eight. We use WikiText-103-raw-v1's training split~\citep{wikitext}, partitioned into disjoint FIT and TEST blocks for the probes. Each 129-token block contributes one experimental token at position 127 of its 128-token context. Holdout is relative to probe fitting; backbone pretraining overlap is unassessed. Full model identifiers, revisions, and protocols are in the linked code archive.

EIPC uses 1,024 FIT and 1,024 TEST samples; EPD and its post-hoc RMO analysis share 512/256 samples. EPD excludes the earlier EIPC and XEC blocks. JetMoE replication RMC uses 512/256 samples; NHD reuses those exact captures and split. All scalers and principal component analysis (PCA) transforms are fitted on FIT. Linear probes use ridge regression with $\alpha=1$.

For the nonlinear comparison, write $R=S_{l-1}$, $H=S_{l-4:l-2}$, and $Y=g_l$. We compare predictors of $Y$ from $R$ and from $[H;R]$. The recent state has eight raw binary coordinates and older history has 24; NHD uses no PCA. Both MLPs have one 32-unit GELU hidden layer. A 77-unit recent-only MLP matches the history model's parameter count to within three parameters (1,317 versus 1,320). Training uses full-batch AdamW, learning rate $10^{-3}$, weight decay $10^{-4}$, at most 300 epochs, and validation patience 30. A fixed 384/128 split within FIT selects checkpoints; initialization seeds are 42, 123, and 2026.

\begin{table*}[t]
\centering\small
\caption{\textbf{History improves prediction beyond the previous layer.} Top: held-out $R^2$ for nested history windows; $k$ includes the immediately preceding layer. Middle: OLMoE gains over $k=1$ and between consecutive windows. Bottom: JetMoE gains, with 95\% paired bootstrap intervals for the preregistered $k=4$ comparison (10,000 resamples). Dashes denote unavailable windows.}
\label{tab:history}
\begin{tabular}{lrrrrr}
\toprule
Model / target & $k=1$ & $k=2$ & $k=4$ & $k=8$ & $k=11$\\
\midrule
OLMoE / L12 & 0.59879 & 0.62320 & 0.64705 & 0.66265 & 0.66544 \\
JetMoE / L12 & 0.20924 & 0.34982 & 0.35199 & 0.38275 & -- \\
JetMoE / L20 & 0.13898 & 0.24936 & 0.34425 & 0.35463 & -- \\
\midrule
OLMoE cumulative gain & -- & 0.02441 & 0.04826 & 0.06387 & 0.06665 \\
OLMoE stepwise gain & -- & 0.02441 & 0.02385 & 0.01561 & 0.00279 \\
\bottomrule
\end{tabular}
\par\smallskip
\begin{tabular}{lrrrr}
\toprule
JetMoE target & $\Delta_2$ & $\Delta_4$ & $\Delta_8$ & 95\% CI for $\Delta_4$\\
\midrule
L12 & 0.14057 & 0.14275 & 0.17351 & [0.10998, 0.17916] \\
L20 & 0.11038 & 0.20528 & 0.21565 & [0.16891, 0.24396] \\
\bottomrule
\end{tabular}
\end{table*}

To examine the remaining variation directly, separate MLPs predict $Y$ and $H$ from $R$. Their residuals are
\begin{equation}
\epsilon_Y=Y-f(R),\qquad \epsilon_H=H-h(R),
\end{equation}
where $f$ and $h$ are the respective recent-state predictors. Two-fold cross-fitting constructs FIT residuals: each 256-sample fold is predicted by models fitted on the other fold, with validation internal to that training fold. A ridge probe learns $\epsilon_H\mapsto\epsilon_Y$. TEST residualizers use the original FIT training/validation protocol; all residualizers use seed 42. A fixed permutation of TEST history residuals provides a descriptive control. This measures residual predictability relative to the fitted decoder family.

\section{Evidence}
\subsection{Expert selection carries the accessible signal}
EIPC compares three histories of selected-expert contributions: their fused sum, vectors retained in expert-identity slots, and the same vectors with identity slots shuffled. Contributions are projected from 2,048 to 32 dimensions before each full history receives a common 64-component PCA budget. Identity preservation improves $\rsq$ over fusion by \metric{EipcEightA} and \metric{EipcTwelveA}; its mean gain is \metric{EipcMeanA} (Table~\ref{tab:provenance}). The shuffled comparison measures accessibility under compression, since the fused sum remains recoverable before compression.

EPD separates binary selection paths, rank-ordered contribution content, fused outputs, and full provenance under a 32-component budget. The path alone reaches $\rsq=\metric{EpdId}$, while content reaches \metric{EpdContent}. Adding content to the path gives \metric{EpdPathContent}. Single-layer path prediction generally strengthens toward the target, from \metric{EpdFirst} to \metric{EpdLast}, while content remains close to zero. The full-provenance representation retains 15.18\% of its variance under PCA, versus 52.17\% for the path. These results identify selection identity as the strongest accessible predictor and motivate testing how much of its signal comes from the previous layer.

\begin{table*}[t]
\centering\small
\caption{\textbf{Nonlinear decoding preserves the historical advantage (NHD).} Raw-state ridge scores and MLP scores for all three initialization seeds. $A$ is the recent-only MLP gain over recent-only ridge; $B$ is the history MLP gain over the recent-only MLP; $B_{\mathrm{matched}}$ uses the parameter-matched recent-only MLP. Residual scores predict $\epsilon_Y$ from $\epsilon_H$ and use a different target from the joint prediction scores. The permutation control shuffles TEST history residuals with the fitted residual probe held fixed.}
\label{tab:nonlinear}
\begin{tabular}{llrrrrr}
\toprule
Target & Decoder / input & Parameters & Seed 42 & Seed 123 & Seed 2026 & Mean / fixed\\
\midrule
L12 & Ridge / $R$ & -- & -- & -- & -- & 0.20972 \\
L12 & Ridge / $[H;R]$ & -- & -- & -- & -- & 0.37188 \\
L12 & MLP / $R$ & 552 & 0.20871 & 0.21206 & 0.21257 & 0.21111 \\
L12 & Matched MLP / $R$ & 1,317 & 0.22051 & 0.21742 & 0.21633 & 0.21809 \\
L12 & MLP / $[H;R]$ & 1,320 & 0.38018 & 0.38677 & 0.38051 & 0.38249 \\
\midrule
L20 & Ridge / $R$ & -- & -- & -- & -- & 0.13948 \\
L20 & Ridge / $[H;R]$ & -- & -- & -- & -- & 0.35382 \\
L20 & MLP / $R$ & 552 & 0.14678 & 0.15214 & 0.14760 & 0.14884 \\
L20 & Matched MLP / $R$ & 1,317 & 0.14602 & 0.15210 & 0.14688 & 0.14834 \\
L20 & MLP / $[H;R]$ & 1,320 & 0.36819 & 0.36339 & 0.37078 & 0.36745 \\
\bottomrule
\end{tabular}
\par\smallskip
\begin{tabular}{lrrrrrr}
\toprule
Target & Linear gain & $A$ & $B$ & $B_{\mathrm{matched}}$ & Residual $R^2$ & Permuted $R^2$\\
\midrule
L12 & 0.16216 & 0.00139 & 0.17137 & 0.16440 & 0.20549 & -0.33325 \\
L20 & 0.21434 & 0.00936 & 0.21861 & 0.21912 & 0.23556 & -0.32149 \\
\bottomrule
\end{tabular}
\end{table*}

\subsection{Older selections add value across architectures}
RMO holds the OLMoE Layer-11 representation fixed while extending history toward earlier layers. The recent state and each older-history window receive separate 16-component PCA transforms. Every ridge input has 32 coordinates; the baseline pads the older block with zeros. Prediction of Layer 12 increases from \metric{RmoOne} to \metric{RmoEleven} (Table~\ref{tab:history}). Gains arrive most strongly in the nearby layers; extending from eight to eleven historical layers adds \metric{RmoStepEleven}. This exploratory analysis uses the already-observed EPD samples.

RMC tests the same comparison in JetMoE at preregistered target Layers 12 and 20. The recent eight-dimensional selection vector remains unchanged; older history receives eight PCA components, yielding a 16-dimensional input for every history window. The frozen primary comparison is $k=4$ against $k=1$. Its $\Delta_4$ values are \metric{RmcTwelveDelta} and \metric{RmcTwentyDelta}, and both paired bootstrap intervals exclude zero. All 10,000 resamples give positive differences at both targets (seed 314159). Both targets satisfy the frozen rule: a positive baseline, $\Delta_4\geq0.02$, and an interval lower bound above zero.

JetMoE has separate attention and MLP mixtures. The captured MLP router was identified by object identity and matched direct recomputation from its input exactly (maximum absolute logit difference $0$); it differed from the attention router in every block. Thus the replication concerns MLP expert selection throughout.

\subsection{The gain survives nonlinear decoding}
One explanation for a linear history gain is that older selections expose structure already present nonlinearly in the recent state. NHD tests that explanation on the frozen JetMoE captures with uncompressed selection vectors. Recent-only MLPs improve over ridge by \metric{NhdTwelveA} and \metric{NhdTwentyA}, whereas adding history improves the MLP by \metric{NhdTwelveB} and \metric{NhdTwentyB} (Table~\ref{tab:nonlinear}). The corresponding parameter-matched gains are \metric{NhdTwelveMatched} and \metric{NhdTwentyMatched}. Every initialization gives a positive history gain against both recent-only controls.

Cross-fitted history residuals also predict the target residuals: $\rsq=\metric{NhdTwelveResidual}$ and $\metric{NhdTwentyResidual}$. Permuting TEST history residuals reduces these scores to \metric{NhdTwelvePermutation} and \metric{NhdTwentyPermutation}. The joint and residual results agree: the tested nonlinear recent-state decoders leave substantial predictive structure accessible from older selections. Both targets receive the frozen descriptive classification \texttt{RESIDUAL-HISTORY-VALUE}.

\section{Discussion and Scope}
Expert trajectories predict routing, older selections add value given the latest selection, and stronger recent-only decoders retain the gap. Across the OLMoE exploration and JetMoE replication, the local selection pattern is an incomplete predictive summary under these probes.

The state definition matters. $R$ records selected identities, rather than the full hidden representation or the preceding router's continuous logits. The findings establish model-relative residual predictive value; they do not establish causal memory, a formal Markov order, or an information-theoretic proof that $H$ contains information mathematically absent from $R$. History comparisons cover two models, three target/model combinations, one corpus, and one position per block. NHD reuses the replication samples. The fixed PCA budget changes the representation as the history window grows, so these curves do not identify a memory-decay law. Mapping effective historical depth across models and layers remains future work.

\paragraph{Prediction and actionability.}
The closed XEC experiment evaluated a five-action Layer-12 routing policy on OLMoE (2,048 training, 512 validation, 1,024 TEST samples; three seeds). Mean next-token NLLs were 2.692297 for native routing, 2.690444 for a current-only policy, 2.689735 for a fused cache, 2.690791 for an expert cache, and 2.647312 for the five-action oracle. Expert-cache differences versus native, current-only, and fused routing were respectively $-0.001506$, $+0.000347$, and $+0.001056$; their 95\% paired intervals were $[-0.005743,0.002760]$, $[-0.000761,0.001670]$, and $[-0.000023,0.002156]$. The experiment did not demonstrate reliable NLL improvement. Validation selected the untrained initialization for every learned policy, which bounds the interpretation to this training and decision setup. The result distinguishes measurable historical structure from its successful use in routing control.

\section{Conclusion}
Earlier expert selections improve held-out routing prediction beyond the previous-layer baseline in OLMoE and JetMoE. Nonlinear and parameter-matched decoders preserve the gain, and residualized history predicts the residual target. These results identify predictive structure in routing trajectories that the tested recent-state summaries leave unexplained.

\clearpage
\balance
\renewcommand{\bibpreamble}{\interlinepenalty=10000}
\bibliographystyle{unsrtnat}
\bibliography{main}
\end{document}